\documentclass{article}
\usepackage{graphicx}

\usepackage{report}
\usepackage{wrapfig}
\usepackage{multicol}
\usepackage{tabularx}
\usepackage{aliascnt}
\usepackage{adjustbox}
\usepackage{makecell}
\usepackage{natbib}
\usepackage{listings}
\usepackage{makecell}
\usepackage{fontawesome5}
\usepackage{soul}
\usepackage[utf8]{inputenc} 
\usepackage[T1]{fontenc}    
\usepackage{hyperref}       
\usepackage{url}            
\usepackage{booktabs}       
\usepackage{amsfonts}       
\usepackage{fancyhdr}       

\usepackage{nicefrac}       
\usepackage{microtype}      
\usepackage[svgnames]{xcolor}         
\usepackage{graphicx}
\usepackage{pifont}
\usepackage{multirow}
\usepackage{CJKutf8}
\definecolor{darkmagenta}{rgb}{0.56, 0.0, 1.0}
\definecolor{softyellow}{rgb}{1.0, 0.92, 0.3} 
\definecolor{LightAquamarine}{rgb}{0.75, 1.0, 0.8} 
\definecolor{FireBrick}{RGB}{178,34,34}
\definecolor{MediumPurple}{RGB}{147,112,219}

\definecolor{uclablue}{rgb}{0.15, 0.45, 0.68}
\hypersetup{
    breaklinks,
    colorlinks=true,
    citecolor={darkmagenta},
    linkcolor={uclablue},
    urlcolor={uclablue}
}
\usepackage{float}
\usepackage{subcaption}
\usepackage{placeins}
\usepackage{lipsum}
\usepackage{tcolorbox}
\tcbuselibrary{skins}
\usepackage{amsmath}
\usepackage{amssymb}
\usepackage{utfsym}
\usepackage{xspace}
\usepackage{enumitem}
\usepackage{multirow} 
\tcbuselibrary{breakable}
\usepackage{colortbl}

\usepackage{tcolorbox}
\usepackage{transparent}
\usepackage{amssymb}
\usepackage{bbding}
\usepackage{soul}
\definecolor{uclablue}{rgb}{0.15, 0.45, 0.68}
\definecolor{mygreen}{rgb}{0.286, 0.616, 0.369}
\definecolor{mygreen2}{rgb}{0.839, 0.996, 0.816}
\definecolor{myyellow2}{rgb}{0.996, 0.980, 0.863}
\definecolor{mycolor_blue}{HTML}{E7EFFA}
\definecolor{mycolor_green}{HTML}{E6F8E0}

\definecolor{njuPurple}{RGB}{220,205,230}     
\definecolor{njuPurpleLight}{RGB}{250,245,252}   
\newtcolorbox{abstractbox}{
    colback=njuPurpleLight,   
    colframe=njuPurple,       
    boxrule=1pt,              
    arc=4mm,                  
    left=8pt,                 
    right=8pt,                
    top=8pt,                  
    bottom=8pt,               
    opacityback=0.95
}

\definecolor{instrback}{RGB}{199,228,179} 
\definecolor{instrframe}{RGB}{4,4,4}

\definecolor{qback}{RGB}{172,214,142}     
\definecolor{qframe}{RGB}{4,4,4}    

\definecolor{aback}{RGB}{255,255,224}     
\definecolor{aframe}{RGB}{4,4,4}    
\definecolor{lightblue}{RGB}{199,228,179}
\definecolor{darkblue}{RGB}{172,214,142}
    
\newtcolorbox{instrbox}[1][]{%
	breakable,
	colback=instrback,
	colframe=instrframe,
	arc=2mm,
	boxrule=0.4pt,
	left=6pt, right=6pt, top=6pt, bottom=6pt,
	#1
}

\newtcolorbox{qabox}[1][]{%
	breakable,
	colback=qback,
	colframe=qframe,
	arc=2mm,
	boxrule=0.4pt,
	left=6pt, right=6pt, top=6pt, bottom=6pt,
	#1
}

\newtcolorbox{aabox}[1][]{%
	breakable,
	colback=aback,
	colframe=aframe,
	arc=2mm,
	boxrule=0.4pt,
	left=6pt, right=6pt, top=6pt, bottom=6pt,
	#1
}

\definecolor{myLightYellow}{RGB}{255,255, 224} 
\definecolor{myDeepYellow}{RGB}{126, 140, 92}
\definecolor{myDeepGreen}{RGB}{80, 128, 70}

\definecolor{boxgreen}{RGB}{230, 245, 233}
\definecolor{boxframegreen}{RGB}{70, 150, 80}
\newtcolorbox{checkblock}{
    colback=boxgreen,           
    colframe=boxframegreen,     
    fonttitle=\bfseries,        
    coltitle=black,             
    boxrule=0.8pt,              
    breakable,                  
    pad at break=1mm,           
    toptitle=0.2mm,
    bottomtitle=0.2mm,
    left=0.2mm,    
    right=0.2mm,   
    top=0.2mm,     
    bottom=0.2mm   
}

\newtcolorbox{contentblock-2}[1]{
    fontupper = \small,
    breakable,
    colback=myLightYellow,   
    colframe=myDeepGreen,   
    title={#1},              
    boxrule=1pt,             
    left=0.2mm,    
    right=0.2mm,   
    top=0.2mm,     
    bottom=0.2mm   
}

\newtcolorbox{contentblock}[1]{
    fontupper = \scriptsize,
    breakable,
    colback=myLightYellow,   
    colframe=myDeepYellow,   
    title={#1},              
    boxrule=1pt,             
    left=0.2mm,    
    right=0.2mm,   
    top=0.2mm,     
    bottom=0.2mm   
}

\newtcolorbox{rulebox}[1]{
      colback=lightblue,     
      colframe=darkblue,     
      boxrule=0.4pt,
      fonttitle=\bfseries,
      title={#1},
      sharp corners,
      coltitle=black,
      enhanced
}

\newcolumntype{C}[1]{>{\centering\arraybackslash}m{#1}}
\newcolumntype{L}[1]{>{\raggedright\arraybackslash}m{#1}}
\newcommand*\samethanks[1][\v alue{footnote}]

\title{
MT-Video-Bench: A Holistic Video Understanding Benchmark for Evaluating Multimodal LLMs in Multi-Turn Dialogues
}

\author{
\textbf{Yaning Pan$^{1,*}$},
\textbf{Qianqian Xie$^{2,*}$},
\textbf{Guohui Zhang$^{3}$},
\textbf{Zekun Wang$^4$},
\textbf{Yongqian Wen$^2$},
\textbf{Yuanxing Zhang$^4$},
\textbf{Haoxuan Hu$^2$},
\textbf{Zhiyu Pan$^2$},
\textbf{Yibing Huang$^2$},
\textbf{Zhidong Gan$^2$},
\textbf{Yonghong Lin$^2$},
\textbf{An Ping$^2$},
\textbf{Shihao Li$^2$},
\textbf{Yanghai Wang$^2$},
\textbf{Tianhao Peng$^2$},
\textbf{Jiaheng Liu$^{2,\dagger}$}
\\
\vspace{4mm}
{\normalsize  $^1$Fudan University}, \quad
{\normalsize  $^2$NJU-LINK Team, Nanjing University}, \\
{\normalsize  $^3$University of Science and Technology of China}, \quad
{\normalsize  $^4$Kuaishou Technology},
\vspace{2mm}
\texttt{ynpan24@m.fudan.edu.cn}
\quad\quad\quad
\texttt{liujiaheng@nju.edu.cn} \\
}

\begin{document}

\maketitle
\let\oldthefootnote\thefootnote

\let\thefootnote\relax\footnotetext{*~Equal Contribution. $^\dagger$~Corresponding Author.}
\let\thefootnote\oldthefootnote

\begin{abstractbox}
\begin{center}
\textbf{\Large Abstract}
\end{center}

The recent development of Multimodal Large Language Models (MLLMs) has significantly advanced AI's ability to understand visual modalities. However, existing evaluation benchmarks remain limited to single-turn question answering, overlooking the complexity of multi-turn dialogues in real-world scenarios. To bridge this gap, we introduce \textbf{MT-Video-Bench}\footnote{\url{https://github.com/NJU-LINK/MT-Video-Bench}}, a holistic video understanding benchmark for evaluating MLLMs in multi-turn dialogues. Specifically, our MT-Video-Bench mainly assesses 6 core competencies that focus on perceptivity and interactivity, encompassing 1,000 meticulously curated multi-turn dialogues from diverse domains. These capabilities are rigorously aligned with real-world applications, such as interactive sports analysis and multi-turn video-based intelligent tutoring. With MT-Video-Bench, we extensively evaluate various state-of-the-art open-source and closed-source MLLMs, revealing their significant performance discrepancies and limitations in handling multi-turn video dialogues. The benchmark will be publicly available to foster future research.
\end{abstractbox}

\section{Introduction}

The rapid progress of Multimodal Large Language Models (MLLMs) has markedly advanced AI’s capacity to perceive and reason over visual modalities, especially when integrated with natural language~\citep{rawal2024cinepile, wang2024lvchat, chandrasegaran2024hourvideo, yu2025vrbench}. Recent systems such as Qwen3-VL~\citep{Qwen3-VL}, InternVL3.5~\citep{internvl35}, and Gemini 3~\citep{gemini3} demonstrate impressive performance in single-turn video question answering and long-form video comprehension. Yet, real-world human–AI interaction is rarely confined to single-turn queries. Instead, it typically unfolds as multi-turn dialogues, where users iteratively refine their questions, shift topics, and expect contextually coherent responses grounded in video content. This interactive setting poses unique challenges: models must not only recall and integrate prior dialogue history but also adapt to conversational dynamics, such as handling topic shifting or gracefully refusing unanswerable queries.

Despite these demands, existing video understanding benchmarks~\citep{fu2025video, wang2024lvbench, zhou2025mlvu, yu2025vrbench} predominantly focus on single-turn evaluation, emphasizing factual perception of video content—such as recognizing objects, actions, or temporal relations—while neglecting dialogue-level reasoning. A few recent efforts explore long-context or multi-shot video benchmarks, yet they fall short of capturing the interplay between perceptivity (faithfully interpreting multimodal input) and interactivity (sustaining natural, user-aware conversations). Consequently, the community lacks a rigorous and holistic framework to evaluate MLLMs in realistic multi-turn, video-grounded dialogues.

\begin{figure}[t]
    \centering
    \includegraphics[width=\linewidth]{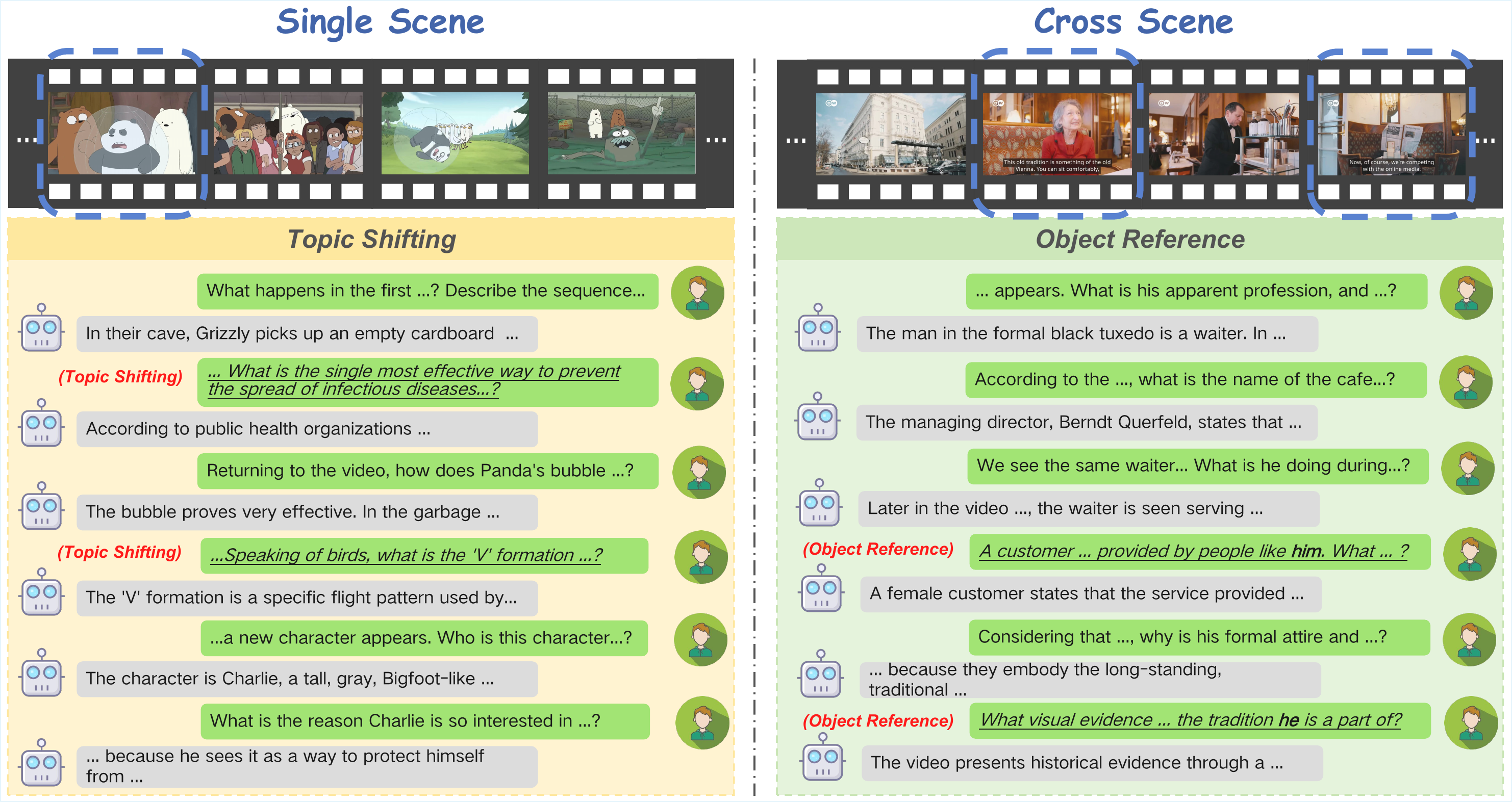}
    \caption{Illustration of multi-turn dialogues under single-scene and cross-scene settings. The evaluated questions corresponding to tasks are marked with underlining, and the scenes involved in the entire multi-turn dialogues are marked with blue dotted boxes.}
    \label{fig:intro}
\end{figure}

To address the above limitations, as shown in Figure~\ref{fig:intro}, we introduce \textbf{MT-Video-Bench}, a holistic benchmark for evaluating MLLMs in multi-turn video dialogues. MT-Video-Bench systematically targets 6 core capabilities spanning perceptivity (object reference, memory recall, and content summary) and interactivity (answer refusal, topic shifting, and proactive interaction). The benchmark comprises 1,000 carefully curated dialogues spanning various video domains, including sports, education, and daily activities. Moreover, unlike prior datasets, MT-Video-Bench emphasizes cross-scene reasoning, long-range dependencies, and interactive adaptability, thereby aligning closely with the demands of real-world applications.

Based on our MT-Video-Bench, we provide a detailed evaluation of open-source and closed-source models, highlighting the limitations and performance discrepancies in different abilities. Specifically, several insightful findings are as follows:
\begin{itemize}
    \item The perceptual and interactive capabilities of MLLMs in multi-turn dialogues still have significant room for improvement. On our MT-Video-Bench, even the high-performance closed-source model Gemini-2.5-Pro achieves only 76.95\% overall accuracy, while most open-sourced MLLMs exhibit accuracies below 60\%, except for the Qwen3-VL series.
    \item Higher scene complexity negatively impacts the accuracy of fine-grained perception. As the number of scenes increases from 1-5 to over 20, even top-tier models like Gemini-2.5-Pro exhibit a steady decline totaling 13\%, highlighting the challenge of long-range temporal dependencies and precise spatio-temporal reasoning within highly fragmented and complex video contexts.
    \item Context reliability becomes increasingly critical as dialogue turns accumulate. The significant gap between golden context and self-predicted settings reveals a ``cascade effect'', where early errors accumulate and lead to dialogue collapse. However, even under an idealized golden context, performance consistently declines as turns grow, highlighting the inherent difficulty of sustaining coherence in extended interactions.

\end{itemize}

To summarize, the contributions of this paper are as follows: We identify the critical gap in evaluating multi-turn video-grounded dialogues and propose MT-Video-Bench, a holistic benchmark that operationalizes this evaluation via six well-defined capabilities across 1,000 dialogues and 5,887 QA pairs. Furthermore, based on extensive experiments, we underscore the challenges and potential directions for improvement in handling and reasoning over multi-turn dialogues, offering a roadmap for future research and development.

\newcommand{\badge}[2]{%
\tikz[baseline=(X.base)] 
\node[draw=#1!60, fill=#1!15, rounded corners=2pt, inner sep=2pt] (X)
{\scriptsize\textbf{#2}};
}

\newcommand{\tOR}{\badge{blue}{OR}}
\newcommand{\tMR}{\badge{teal}{MR}}
\newcommand{\tCS}{\badge{violet}{CS}}
\newcommand{\tAR}{\badge{orange}{AR}}
\newcommand{\tTS}{\badge{brown}{TS}}
\newcommand{\tPI}{\badge{gray}{PI}}

\begin{table*}[th]
    \phantomsection
    \centering
    \resizebox{\textwidth}{!}{
    \begin{tabular}{lcccccccccccc}
        \toprule
        \textbf{Benchmark} & \textbf{\#QAs} & \textbf{Avg. Q/V} & \textbf{Long} & \textbf{Dialogue} & \textbf{\#Turns} & \textbf{Q. Div.} & \textbf{A. Div.} & \textbf{Q. Len.} & \textbf{A. Len.} & \textbf{Task Type (Multi-turn)} \\
        \midrule
        MVBench & 4,000 & 1.00 & \textcolor{red}{\XSolidBrush} & \textcolor{red}{\XSolidBrush} & 1.00 & 23.88 & 6.47 & 14.79 & 4.80 & - \\
        LongVideoBench & 6,678 & 1.77 & \textcolor{red}{\XSolidBrush} & \textcolor{red}{\XSolidBrush} & 1.00 & 46.07 & 12.85 & 51.27 & 11.24  & - \\
        Video-MME & 2,700 & 3.00 & \textcolor{mygreen}{\CheckmarkBold} & \textcolor{red}{\XSolidBrush} & 1.00 & 27.33 & 9.31 & 15.53 & 7.26  & - \\
        LVBENCH & 1,549 & 15.04 & \textcolor{mygreen}{\CheckmarkBold} & \textcolor{red}{\XSolidBrush} & 1.00 & 22.47 & 6.39 & 14.12 & 5.20 & - \\
        MLVU & 3,102 & 1.79 & \textcolor{mygreen}{\CheckmarkBold} & \textcolor{red}{\XSolidBrush} & 1.00 & 26.21 & 11.42 & 17.98 & 16.91 & - \\
        Video-MMLU & 15,746 & 14.78 & \textcolor{red}{\XSolidBrush} & \textcolor{red}{\XSolidBrush} & 1.00 & 16.74 & 16.57 & 11.36 & 10.59  & - \\
        \midrule
        SVBench & 7,374 & 36.87 & \textcolor{red}{\XSolidBrush} & \textcolor{mygreen}{\CheckmarkBold} & 4.29 & 13.67 & 22.55 & 9.97 & 13.49  & \tOR\ \tMR \\
        \textbf{MT-Video-Bench (Ours)} & 5,887 & 43.61 & \textcolor{mygreen}{\CheckmarkBold} & \textcolor{mygreen}{\CheckmarkBold} & 5.89 & 54.18 & 75.84 & 35.51 & 74.54 & \tOR\ \tMR\ \tCS\ \tAR\ \tTS\ \tPI \\
        \bottomrule
    \end{tabular}}
    \caption{Comparison with other benchmarks. \textbf{Avg. Q/V}: the average number of QA pairs per video. \textbf{Long}: whether the average video length is greater than 10 minutes. \textbf{Q. Div.} and \textbf{A. Div.}: the lexical diversity of questions and answers, calculated following \citet{lee2025multiverse}. \textbf{Q. Len.} and \textbf{A. Len.}: the average number of tokens of questions and answers, computed with the LLaMA-3.1-8B tokenizer~\citep{dubey2024llama}. \textbf{OR}: Object Reference. \textbf{MR}: Memory Recall. \textbf{CS}: Content Summary. \textbf{AR}: Answer Refusal. \textbf{TS}: Topic Shifting. \textbf{PI}: Proactive Interaction.}
    \label{tab:benchmark_compare}
\end{table*}

\section{Related Work}
\textbf{Multimodal LLMs.} MLLMs have become a focal point in advancing general-purpose intelligence. By jointly modeling textual and visual modalities, these models capture cross-modal dependencies and enhance semantic reasoning~\citep{zhu2023minigpt, ma2024vista, zhang2024flash, wang2025ross3d, wang2024reconstructive}. Recent advances have further extended MLLMs to the video domain~\citep{li2023videochat, cheng2024videollama, maaz2023video}, enabling video understanding to support grounded dialogue.
For example, Qwen3-VL~\citep{Qwen3-VL} fuses a SigLIP-2-based dynamic-resolution ViT with an interleaved-MRoPE–DeepStack merger to feed multi-layer visual tokens into the Qwen3 LLM, achieving unified 256K-token multimodal reasoning.
InternVL3.5~\citep{internvl35} integrates InternViT as the vision encoder within a ViT-MLP-LLM paradigm, further adopting a visual resolution router and visual consistency learning for cross-modal alignment.

\textbf{Video Benchmarks}. Table~\ref{tab:benchmark_compare} provides a comparative overview of recent progress in video benchmarks. Specifically, Video-MME~\citep{fu2025video} and LongVideoBench~\citep{wu2024longvideobench} utilize multi-choice QA to evaluate model capabilities across various temporal scales. In terms of open-ended evaluation, MVBench~\citep{mvbench} focuses on concise QA tasks to assess fundamental multimodal understanding, while Video-MMLU~\citep{song2025video} emphasizes knowledge-intensive reasoning.
Furthermore, MLVU~\citep{zhou2025mlvu} and LVBENCH~\citep{wang2024lvbench} provide comprehensive assessments specifically tailored for long-video understanding performance. SVBench~\citep{svbench} targets dialogues in real-time streaming videos to assess understanding and reasoning. Despite these advancements, significant gaps remain in multi-turn dialogue, QA complexity and diversity, and comprehensive interactive assessment, motivating the development of MT-Video-Bench.

\section{MT-Video-Bench}
This section provides a detailed presentation of MT-Video-Bench. We first establish the core task definitions in Section \ref{sec:task_definition}, followed by the dataset construction pipeline in Section \ref{sec:dataset_construction}. Sections \ref{sec:dataset_statistics} and \ref{sec:evaluation_method} then detail the benchmark statistics and our evaluation methodology, respectively. 

\subsection{Task Definition}
\label{sec:task_definition}

MT-Video-Bench is designed to comprehensively evaluate the ``Perceptivity'' and ``Interactivity'' of MLLMs in multi-turn video-grounded dialogues, emphasizing contextual coherence, cross-scene video comprehension, and adaptive interactivity.

\textbf{Perceptivity} assesses the model's foundational ability to perceive and integrate information from both the visual video content and the multi-turn conversational context. This capability is essential for accurately understanding user queries and generating contextually grounded responses throughout the dialogue. It includes:

(1) \textbf{Object Reference (OR)} tests the model’s ability to resolve implicit references and pronouns, ensuring accurate mapping to specific video entities or characters.

(2) \textbf{Memory Recall (MR)} measures the capacity to retrieve and integrate information from prior conversational history to maintain reasoning continuity and dialogue coherence.

(3) \textbf{Content Summary (CS)} assesses the effectiveness of condensing video and conversational content into succinct, comprehensive summaries while preserving semantic fidelity.
    
\begin{figure*}[th]
\centering
\includegraphics[width=1\linewidth]{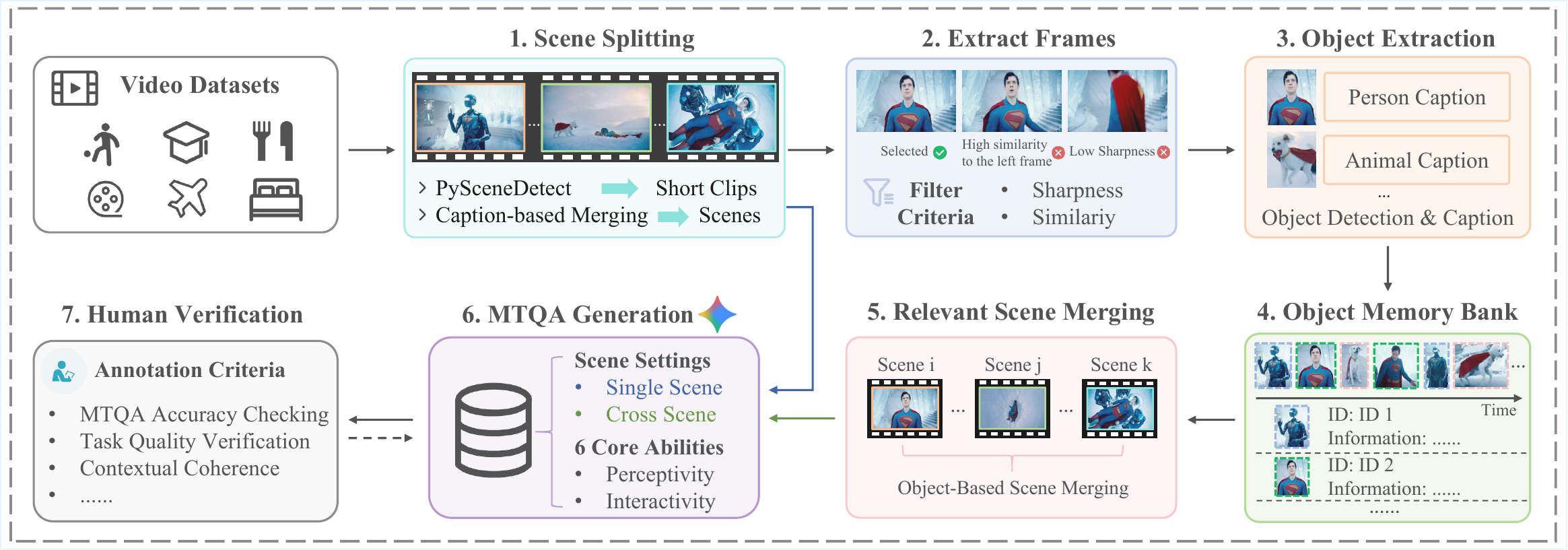}
\caption{An overview of the semi-automatic data construction process of MT-Video-Bench.}
\label{fig: pipeline}
\end{figure*}

\textbf{Interactivity} evaluates the model’s capacity to conduct coherent, adaptive, and user-aware dialogues based on the video content. It focuses on appropriately refusing unanswerable questions, smoothly adapting to topic changes, and proactively maintaining engagement. It includes:

(1) \textbf{Answer Refusal (AR)} assesses the recognition of unanswerable queries and the ability to decline them explicitly without hallucinating.

(2) \textbf{Topic Shifting (TS)} evaluates how the model adapts to user-initiated changes in conversational focus while maintaining coherence and relevance.
    
(3) \textbf{Proactive Interaction (PI)} probes the capacity to sustain engagement via clarifications or novel insights when disinterest is detected, fostering dialogue continuation.

\subsection{Dataset Construction}
\label{sec:dataset_construction}

With the guidance of task definitions, we next collect and annotate high-quality videos for each task. Specifically, we design an annotation pipeline in Figure~\ref{fig: pipeline}, which involves both automated construction and rigorous human verification.

\textbf{Video Collection and Single-Scene Splitting.} The data collection process begins with the manual acquisition of 200 videos from various online platforms like YouTube, followed by rigorous filtering to ensure quality. Subsequently, we employ PySceneDetect\footnote{https://github.com/Breakthrough/PySceneDetect} to divide the videos into shorter clips. Recognizing that these clips are often too brief to represent complete scenes, we then use the Gemini-2.5-Flash to generate descriptive captions for each clip.
Finally, the caption-based clip merging method is iteratively applied twice to combine related clips into a coherent, single-scene video, ensuring a seamless and contextually accurate representation of the scene. These refined single-scene videos serve as the core visual content for the subsequent task of generating single-scene dialogues.

\textbf{Cross-Scene Video Merging.} The generation of cross-scene, multi-turn dialogues necessitates the retrieval and merging of relevant scenes from disparate video segments, which serves as a critical step in creating coherent interactions that span across multiple visual contexts. 
Firstly, frames are extracted from the video at 2 FPS and then filtered based on two criteria: sharpness and similarity to the previous selected frame. The sharpness of each frame is evaluated by the Laplace Operator to ensure that only clear, visually significant frames are retained, improving the overall quality of the selected frames. To avoid redundancy, frames with high similarity to the preceding selected frame are discarded. Specifically, a histogram-based image similarity calculation method is used to compare consecutive frames, excluding those with a similarity score above 0.9. This approach ensures that the selected frames are distinct and capture key moments in the video. 

Following frame selection, object detection is performed using YOLOv11~\citep{khanam2024yolov11}, and each detected object is then annotated with a caption generated by the Gemini-2.5-Flash, providing detailed descriptions for each object. As the video progresses, a dynamic object memory bank is maintained, continuously expanded based on object captions and visual similarities. This memory bank associates unique object IDs with their corresponding attributes, enabling the identification of the same objects across frames. To merge relevant scenes, a retrieval step across scenes is performed to select video segments that share common objects or themes, which are then merged to ensure continuity both thematically and contextually.

\begin{figure*}[t]
\centering
\includegraphics[width=1.0\linewidth]{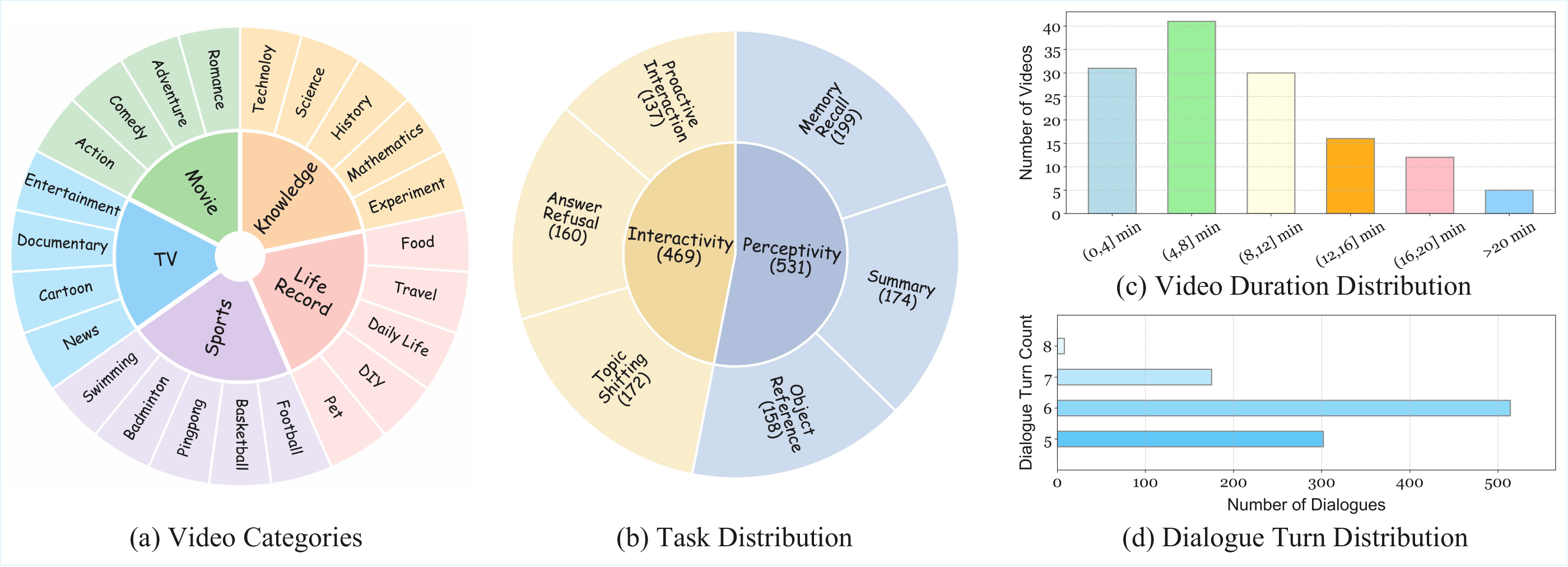} 
\caption{Overview of MT-Video-Bench. (a) Video Categories. MT-Video-Bench includes videos spanning 5 major categories, ensuring diverse topical coverage. (b) Task Distribution. MT-Video-Bench consists of a total of 6 tasks with a relatively balanced distribution. (c) Video Duration Distribution. MT-Video-Bench includes both long and short videos. (d) Dialogue Turn Distribution. Multi-turn dialogues in MT-Video-Bench involve 5 to 8 rounds.}
\label{fig: category}
\end{figure*}

\textbf{Multi-Turn Dialogues Generation.} This process employs the Gemini-2.5-Pro—which yielded the highest generation quality in our preliminary experiments—to automate the generation of both single-scene and cross-scene multi-turn dialogues, based on the 6 evaluation tasks defined earlier. For each video, we generate multiple multi-turn dialogues, each corresponding to different scenes. To determine the most appropriate task for each scene, we prompt MLLMs to evaluate the scene’s capabilities, scoring them on a scale from 1 to 6. Only those tasks that receive a score of 5 or 6 are selected for dialogue generation. For multi-turn dialogues spanning multiple scenes, we specifically adopt an object-centered approach for cross-scene question design since objects often serve as the central element around which events unfold. This approach emphasizes the continuity and relationships of objects across scenes, enabling the generation of dialogues that are both contextually consistent and thematically coherent.

\textbf{Human Verification.} Following the automated phase, we conduct rigorous human verification to ensure data quality. First, we examine whether each dialogue correctly aligns with its intended capability dimension. For instance, answer refusal tasks must test the recognition of absent events, while object reference must involve effective pronoun disambiguation. Second, annotators verify that each question-answer pair is factually grounded in the video and free from linguistic ambiguities. Additionally, we assess contextual coherence to ensure that each dialogue turn follows logically from the previous conversation history, maintaining a consistent narrative thread. To guarantee the reliability of these annotations, each dialogue is independently reviewed by two annotators. In cases of disagreement, a senior annotator facilitates a discussion to reach a final consensus. This process yielded the final 1,000 high-quality multi-turn dialogues.

\subsection{Dataset statistics}
\label{sec:dataset_statistics}

Figure \ref{fig: category} presents the statistics of MT-Video-Bench. It covers a broad range of topics across 5 main categories: Movie, TV, Sports, Knowledge, and Life Record, each with multiple sub-topics, ensuring a diverse and balanced data distribution. With a total of 1,000 multi-turn dialogues, the data distribution across the 6 primary tasks in MT-Video-Bench is relatively balanced, as shown in Figure \ref{fig: category} (b). Furthermore, our dataset features videos of varying lengths, with most being under 15 minutes and a small proportion exceeding 15 minutes, thereby ensuring coverage of both short and long videos. The number of dialogue turns typically ranges from 5 to 8, with an average of 5.89 turns per dialogue.

\begin{table*}[th]
\centering
\resizebox{1.0\textwidth}{!}{
\begin{tabular}{l >{\centering\arraybackslash}p{1.0cm} >{\centering\arraybackslash}p{0.7cm} >{\centering\arraybackslash}p{0.7cm} >{\centering\arraybackslash}p{0.7cm} >{\centering\arraybackslash}p{0.7cm} >{\centering\arraybackslash}p{0.7cm} >{\centering\arraybackslash}p{0.7cm}}
\toprule
\multirow{2}{*}{\centering \textbf{Models}} & \multirow{2}{*}{\centering \textbf{Overall}} & \multicolumn{3}{c}{\textbf{Perceptivity}} & \multicolumn{3}{c}{\textbf{Interactivity}}\\
\cmidrule(lr){3-5} \cmidrule(lr){6-8}
 &  & OR & MR & CS & AR & TS & PI \\
\midrule
\multicolumn{8}{c}{\textit{Closed-Sourced Models}} \\
\midrule
Gemini-2.5-Pro \citep{gemini2.5pro} & 
\textbf{76.95} & \textbf{71.63} & \textbf{72.45} & \textbf{93.71} & \textbf{57.74} & \textbf{89.67} & \textbf{76.50} \\
Gemini-2.5-Flash \citep{gemini2.5pro}  & \underline{69.90} & \underline{64.07} & \underline{67.23} & 92.45 & 47.50 & \underline{83.17} & 64.98 \\
Doubao-Seed-1.6-250615 \citep{Doubao-Seed-1.6} & 67.40 & 53.82 & 57.20 & \underline{93.21} & \underline{55.57} & 81.30 & 63.30 \\
\midrule
\multicolumn{8}{c}{\textit{Open-Sourced Models}} \\
\hline
\rowcolor{gray!20}
\multicolumn{8}{l}{\textit{Model Size $>$ 8B}} \\
Qwen3-VL-32B-Thinking \citep{bai2025qwen2}  & 68.57 & 58.50 & 59.11 & \underline{93.21} & 52.93 & 81.72 & 65.93 \\
Qwen3-VL-32B-Instruct \citep{bai2025qwen2} & 67.84 & 55.57 & 59.94 & 91.40 & 50.48 & 82.88 & 66.80 \\
InternVL3.5-38B (Think) \citep{wang2025internvl3}  & 58.50 & 52.03 & 48.51 & 77.39 & 37.84 & 67.87 & 67.34 \\
InternVL3.5-38B (No Think) \citep{wang2025internvl3}  & 53.74 & 47.34 & 42.85 & 66.99 & 34.92 & 62.23 & 68.09 \\
\hline
\rowcolor{gray!20}
\multicolumn{8}{l}{\textit{4B $<$ Model Size $\leq$ 8B}} \\
Qwen3-VL-8B-Thinking \citep{bai2025qwen2} & 65.98 & 54.78 & 58.21 & 90.97 & 45.35 & 78.01 & 68.59 \\
Qwen3-VL-8B-Instruct \citep{bai2025qwen2} & 62.96 & 53.20 & 54.48 & 90.25 & 44.87 & 73.45 & 61.49 \\
InternVL3.5-8B (Think) \citep{wang2025internvl3}  & 58.36 & 50.70 & 51.58 & 73.83 & 38.83 & 68.74 & 66.47 \\
MiniCPM-V4.5 \citep{yao2024minicpm} & 56.84 & 54.53 & 48.57 & 77.86 & 39.46 & 61.29 & 59.35 \\
Qwen2.5-VL-7B \citep{bai2025qwen2}  & 54.92 & 47.59 & 44.57 & 77.26 & 43.16 & 60.38 & 56.57 \\
LLaVA-Video-7B \citep{zhang2025llava}  & 53.76 & 43.00 & 41.75 & 82.38 & 39.07 & 59.38 & 56.99 \\
LLaVA-OneVision-7B \citep{li2024llava}  & 52.97 & 42.66 & 39.97 & 82.11 & 36.02 & 58.98 & 58.09 \\
InternVL3.5-8B (No Think) \citep{wang2025internvl3} & 52.20 & 45.61 & 39.01 & 63.59 & 33.01 & 60.52 & \underline{71.46} \\
MiniCPM-o \citep{yao2024minicpm} & 50.39 & 46.66 & 39.79 & 74.80 & 27.20 & 56.26 & 57.64 \\
VideoChat-Flash-7B \citep{li2024videochat}  & 46.02 & 41.53 & 38.28 & 69.16 & 24.67 & 51.37 & 51.12 \\
LLaVA-NeXT-Video-7B \citep{llava-next}  & 45.66 & 38.87 & 32.45 & 79.09 & 23.79 & 42.57 & 57.16 \\
VideoLLaMA3-7B \citep{bai2025qwen2} & 38.10 & 41.59 & 31.15 & 47.88 & 29.26 & 35.53 & 43.20 \\
InternVideo2.5-8B \citep{wang2025internvideo2}  & 38.09 & 31.64 & 34.61 & 41.56 & 25.55 & 50.54 & 44.66 \\

\hline
\rowcolor{gray!20}
\multicolumn{8}{l}{\textit{Model Size $\leq$ 4B}} \\
Qwen3-VL-4B-Thinking \citep{bai2025qwen2}  & 62.96 & 52.18 & 55.26 & 91.37 & 41.67 & 72.00 & 65.31 \\
Qwen3-VL-4B-Instruct \citep{bai2025qwen2}  & 59.15 & 49.29 & 46.74 & 83.66 & 52.99 & 65.34 & 56.88 \\
InternVL3.5-4B (Think) \citep{wang2025internvl3}  & 54.58 & 47.80 & 45.02 & 70.52 & 31.82 & 62.39 & 69.96 \\
InternVL3.5-4B (No Think) \citep{wang2025internvl3}  & 49.50 & 41.87 & 40.31 & 60.56 & 28.14 & 55.71 & 70.43 \\
\bottomrule
\end{tabular}
}
\caption{Evaluation results on MT-Video-Bench. \textbf{OR}: Object Reference. \textbf{MR}: Memory Recall. \textbf{CS}: Content Summary. \textbf{AR}: Answer Refusal. \textbf{TS}: Topic Shifting. \textbf{PI}: Proactive Interaction. The best and second-best performances are highlighted in \textbf{bold} and \underline{underlined}, respectively.}
\label{tab:mainresults}
\end{table*}

\subsection{Evaluation Method}
\label{sec:evaluation_method}

In multi-turn dialogues, each new turn depends on the interactions between users and assistants in previous turns. This dynamic is particularly crucial in tasks that involve high interactivity, such as proactive interactions. Therefore, we follow the multi-turn dialogue evaluation setup used in LLMs~\citep{bai2024mt}, leveraging our meticulously curated dataset as the golden context for dialogue history, rather than relying on potentially inconsistent self-predicted context from MLLMs. 

Following the checklist-based approach \citep{liu2024playground,lee2025multiverse}, we use Gemini-2.5-Flash to construct 5 yes/no questions per QA pair, designed to assess response accuracy and task-specific performance.
Then, manual validation is employed to revise unqualified checklists. During the evaluation process, an automated evaluator provides judgments for each checklist question based on the model-generated answers. The score for each multi-turn dialogue is calculated as the accuracy, representing the proportion of correct items across its checklists. The final performance for each task is then calculated by averaging the scores of all multi-turn dialogues under this task.

\section{Experiments}

\subsection{Experimental Settings}

\textbf{Evaluated Models.} For closed-source models, we evaluate 3 popular models including Gemini-2.5-Pro~\citep{gemini2.5pro}, Gemini-2.5-Flash~\citep{gemini2.5pro}, and Doubao-Seed-1.6~\citep{Doubao-Seed-1.6}. For open-source models, we select 21 representative MLLMs, including Qwen3-VL series~\citep{bai2025qwen2}, Qwen2.5-VL series~\citep{bai2025qwen2}, InternVL3.5 series~\citep{wang2025internvl3}, MiniCPM series~\citep{yao2024minicpm}, LLaVA-Onevision-7B~\citep{llavaOnevision}, InterVideo2.5-8B~\citep{internvideo}, LLaVA-Video-7B~\citep{llavaVideo}, LLaVA-NeXT-Video-7B~\citep{llava-next}, VideoChat-Flash-7B~\citep{videochatflash}, and VideoLlama3-7B~\citep{videollama3}.

\textbf{Input Settings.} For most models, we uniformly sample 128 frames, and each frame is resized so that its longer side is limited to 720 pixels, with the other side scaled proportionally. For the InternVL3.5~\citep{wang2025internvl3} series, we uniformly sample 128 frames, and each frame is resized to 448 × 448, following the model-specific requirements. For the InternVideo2.5-8B~\citep{wang2025internvideo2}, we uniformly sample 128 frames, and each frame is resized to 728 × 728 to adhere to model-specific requirements while minimizing experimental discrepancies.

\begin{figure*}[th]
\centering
\includegraphics[width=1\linewidth]{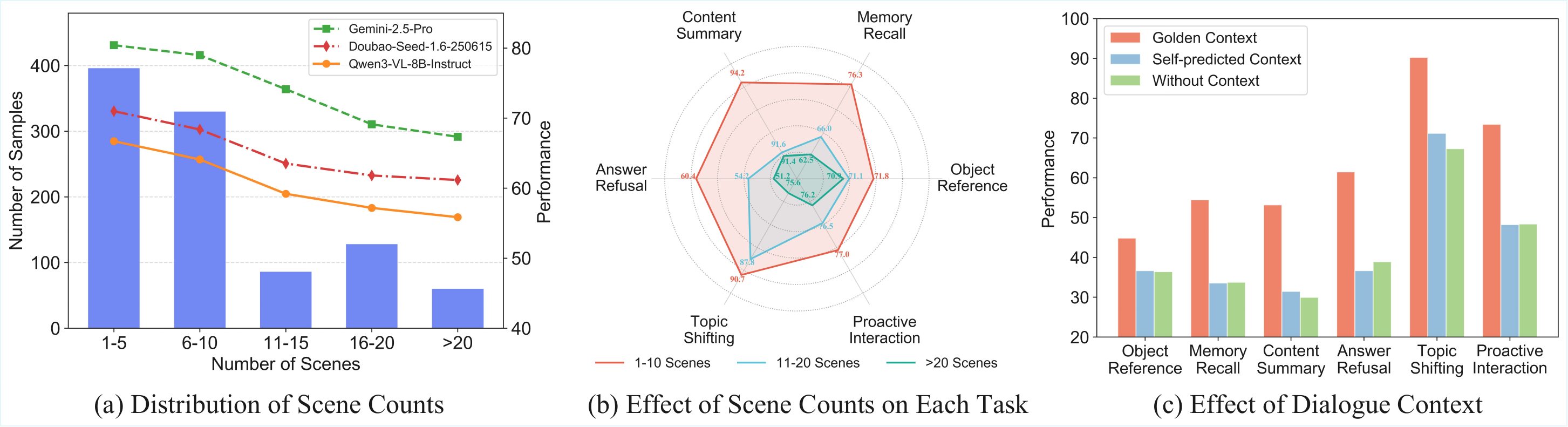}
\caption{The distribution of scene counts involved in each multi-turn dialogue, the effect of scene counts on overall performance and each task performance, and the effect of dialogue context.}
\label{exp1}
\end{figure*}

\subsection{Main Results}

As shown in Table~\ref{tab:mainresults}, we provide the performance of different MLLMs on our MT-Video-Bench, and we have the following key observations:

(1) \textbf{Dominance of closed-source models and shrinking gaps}. Gemini-2.5-Pro maintains a clear lead as the top-performing model across all evaluation metrics. However, Qwen3-VL-32B-Thinking shows highly competitive results, even surpassing Doubao-Seed-1.6 in overall accuracy.

(2) \textbf{Task difficulty variance and universal Bottlenecks}. CS is the most mastered task, with top models frequently exceeding a 90\% success rate. In contrast, AR remains a universal bottleneck, as even the strongest models struggle to effectively recognize knowledge boundaries.

(3) \textbf{Consistent gains from ``Think'' mode}. The ``Think'' or reasoning mode consistently yields substantial performance gains across all model scales. For example, InternVL3.5-38B (Think) achieves significantly higher accuracy than its ``No Think'' counterpart in both perceptivity and interactivity.

\subsection{Further Analysis}
\textbf{Effect of Different Numbers of Scenes.}
Figure \ref{exp1} (a) shows the distribution of scene counts involved in each multi-turn dialogue alongside the performance of three models under different scene counts. We observe that as the number of scenes increases from 1-5 to over 20, even top-tier models like Gemini-2.5-Pro exhibit a consistent performance decline, which suggests that existing MLLMs still struggle to maintain long-range temporal dependencies and precise spatial grounding when faced with highly fragmented and complex video contexts. Specifically for Gemini-2.5-Pro, Figure \ref{exp1} (b) highlights that MR and TS suffer the most pronounced degradation as scene counts increase, while OR and PI remain relatively resilient.

\textbf{Effect of Dialogue Context.}
Figure \ref{exp1} (c) illustrates the impact of three context settings—golden context, self-predicted context, and no context—on the performance of Qwen3-VL-8B-Instruct across diverse tasks. The results reveal that golden context consistently outperforms the other two settings, underscoring the vital role of accurate history in multi-turn reasoning. A significant performance gap exists between golden and self-predicted context, highlighting a severe error propagation issue.

\begin{figure*}[th]
\centering
\includegraphics[width=1\linewidth]{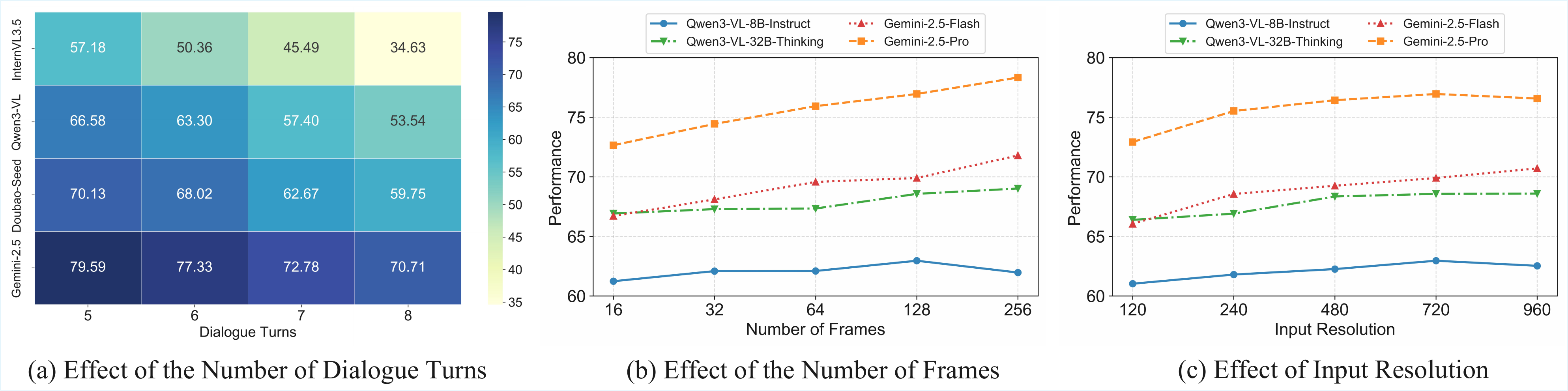}
\caption{Effectiveness of different numbers of dialogue turns and input video parameters on MT-Video-Bench.}
\label{fig: dialogue and video parameters}
\end{figure*}

\textbf{Effect of the Number of Dialogue Turns.}
Figure \ref{fig: dialogue and video parameters} (a) evaluates the models' performance stability as dialogue depth increases from 5 to 8 turns. A universal performance decay is observed across all tested models, confirming that extended multi-turn interactions impose a significant cognitive burden on contextual reasoning. Notably, while Gemini-2.5-Pro maintains the highest baseline and demonstrates relative robustness, InternVL3.5-8B (No Think) exhibits the most pronounced sensitivity to dialogue length, with its performance plunging from 57.18 to 34.63.

\textbf{Effect of Frame and Resolution Settings.} The impact of different input video parameters on performance is illustrated in Figure \ref{fig: dialogue and video parameters} (b) and (c). Fixing the resolution at 720 and varying frame counts (16, 32, 64, 128, 256), the performance generally improves, confirming the benefits of finer temporal sampling. However, a capacity ceiling is observed in the smaller Qwen3-VL-8B-Instruct, where performance saturates at 128 frames and slightly declines at 256 frames. This suggests that for models with limited parameters, excessive temporal information may introduce redundant noise that exceeds their processing bandwidth.
With a constant 128-frame sequence, increasing resolution from 120 to 960, most models peak in performance at 720p, after which they plateau or slightly decline, while Gemini-2.5-Flash continues to show performance enhancement at higher resolutions, attributed to the ability to extract incremental information from ultra-high-resolution visual details.


\begin{wrapfigure}{r}{0.5\textwidth} 
  \centering
  \vspace{-\intextsep}
  \includegraphics[width=\linewidth]{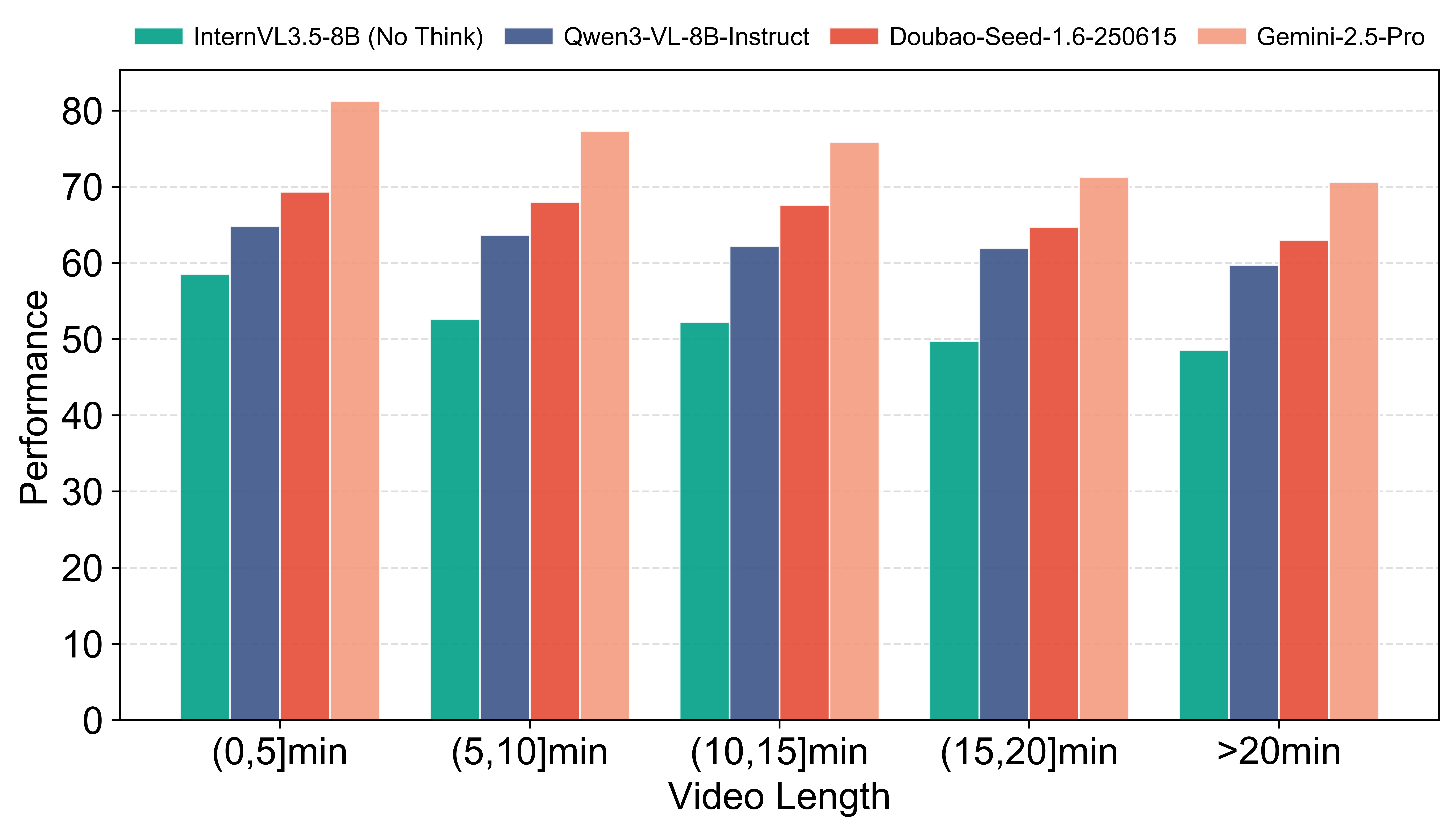}
  \caption{Effect of different video length on MT-Video-Bench evaluated across four representative MLLMs.} 
  \vspace{-\intextsep}
  \vspace{-25pt}
\label{fig: video_length}
\end{wrapfigure}

\textbf{Effect of Different Video Lengths.} Figure \ref{fig: video_length} shows how video length affects performance. While models perform well on shorter clips, a significant drop is observed once the video exceeds 15 minutes. This indicates that as the video grows longer, the increasing complexity makes it much harder for models to maintain high accuracy.

\subsection{Case Study}

Our error analysis identifies three primary failure modes: (i) visual-temporal hallucinations, involving complex entity misattribution and fictitious narrative fabrication; (ii) inter-turn interference, where models confuse previous historical context or struggle with fluid and seamless topic transitions; and (iii) conversational passivity, characterized by a pronounced lack of proactive initiative during prolonged multi-turn engagement. Representative examples illustrating these failure modes are provided in Appendix \ref{sec:more_cases}.


\begin{wraptable}{r}{0.5\textwidth} 
  \centering
  \vspace{-\intextsep}
    \resizebox{0.5\textwidth}{!}{
    \begin{tabular}{cccc}
        \toprule
        \textbf{Model} & \textbf{Spearman} & \textbf{Kendall-Tau} & \textbf{Pearson}   \\
        \midrule
        Gemini-2.5-Flash & \textbf{95.25} & \textbf{87.48} & \textbf{97.01}  \\
        GPT-4o & 93.81 & 83.17 & 95.87 \\
        DeepSeek-V3 & 93.34 & 83.63 & 94.63 \\
        \bottomrule
    \end{tabular}
    }
  \caption{Agreement between automated evaluation and human evaluation across diverse models.} 
  \vspace{-\intextsep}
\label{tab: eval_agreement}
\end{wraptable}

\subsection{Agreement Evaluation}
To assess the alignment between our automated evaluation and human judgment, we randomly sampled 20 dialogues for each task, totaling 120 samples. We recruited three human annotators to evaluate the model responses based on the provided checklists. Following \citep{ye2023flask}, we report the Spearman, Kendall’s Tau, and Pearson correlation to quantify the consistency between the automated assessors (Gemini-2.5-Flash, GPT-4o, and DeepSeek-V3) and human judgment. As shown in Table~\ref{tab: eval_agreement}, the automated evaluations demonstrate a high level of agreement across all metrics, with Gemini-2.5-Flash (our evaluator) achieving the highest consistency. These results indicate that our evaluation pipeline closely aligns with human preferences and provides reliable assessments.

\section{Conclusion}
In this paper, we presented MT-Video-Bench, a holistic benchmark for evaluating MLLMs in multi-turn video dialogues. 
Unlike prior video understanding benchmarks that primarily focus on single-turn factual perception, MT-Video-Bench jointly assesses perceptivity and interactivity through six carefully defined capabilities, covering tasks such as memory recall, topic shifting, and proactive interaction.
Our evaluation of 20 state-of-the-art models provides insightful findings, and we hope our MT-Video-Bench can establish a rigorous foundation for future research, highlighting the need for models that can reason over long contexts while engaging in natural, adaptive conversations.
\clearpage

\bibliography{reference.bbl}
\bibliographystyle{unsrtnat}

\clearpage

\appendix
\section{Details on the Data Generation}

\begin{wrapfigure}{r}{0.4\textwidth} 
  \centering
  \vspace{-\intextsep}
  \includegraphics[width=\linewidth]{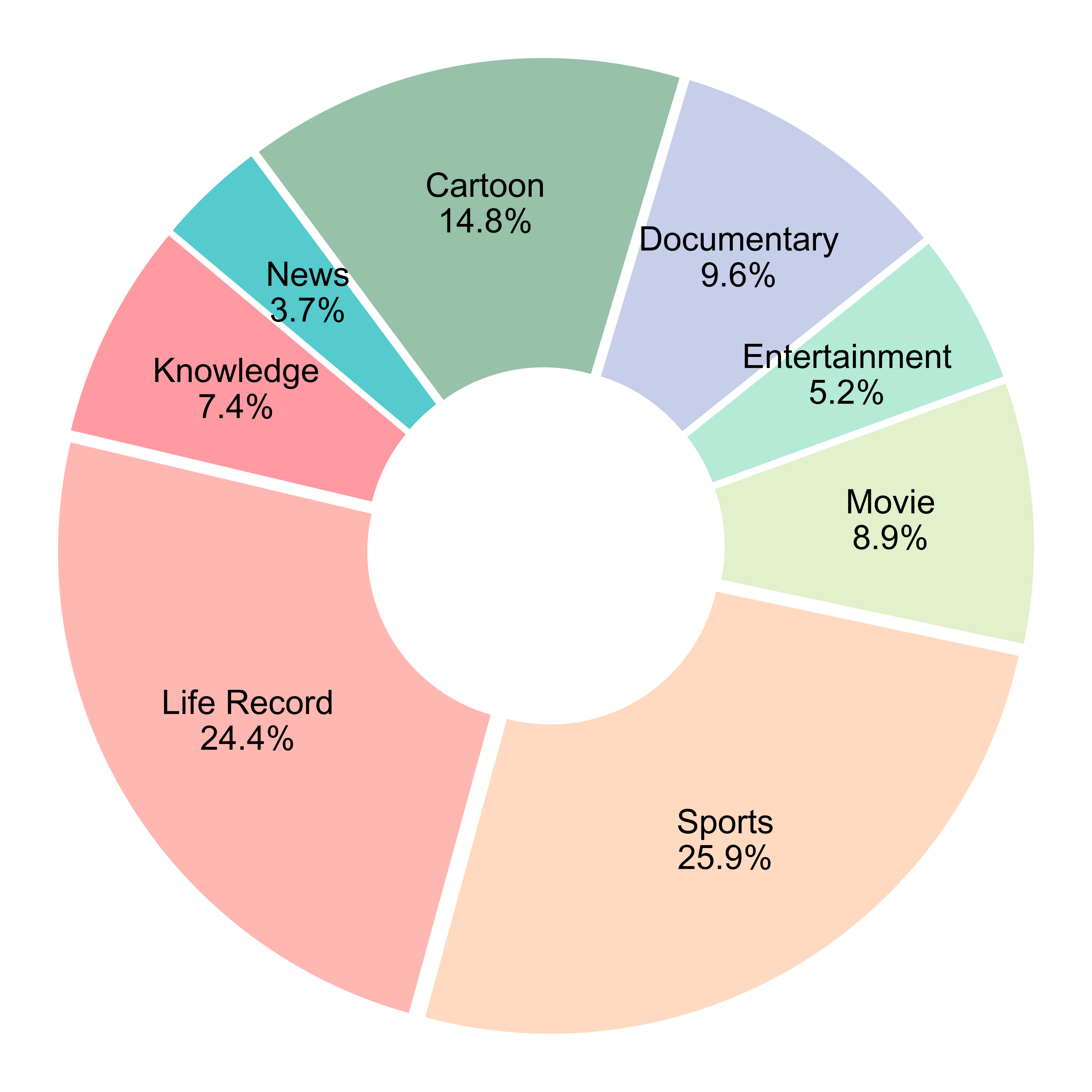}
  \caption{Distribution of video category.} 
  \vspace{-\intextsep}
\label{fig:video_category}
\end{wrapfigure}

\subsection{Distribution of Video Category}
To ensure timeliness and diversity, the video collection consists of English and Chinese content published within the past year. Figure \ref{fig:video_category} illustrates the distribution of video domains within the benchmark, highlighting the diverse range of content categories covered in it.

\subsection{Details on generation prompts}
\label{sec:data_generation}
To ensure the high fidelity and diversity of our data, we designed a structured prompting framework for multi-turn dialogue generation. This framework begins with a unified prompt template that defines the conversational persona and the global context for video-grounded interactions. Following the template, we incorporate task-specific instructions tailored to our 6 evaluation tasks to ensure that the generated data matches our ability and task requirements.
These nuanced requirements guide the generator to move beyond simple question-answering, emphasizing contextual coherence, fine-grained spatio-temporal reasoning, and adaptive interaction that mimic real-world scenarios. Figure~\ref{fig:generate_template} shows the prompt template, while Figures \ref{fig:generate_ts} to \ref{fig:generate_or} show task instructions we utilize for each task.

\section{Details on Evaluation}
This section details our evaluation setup. We first outline the inference settings, followed by the prompts used for checklist generation and the final evaluation process.

\subsection{Inference Settings}
\label{sec:infer_settings}
To ensure a standardized and fair evaluation across diverse MLLM architectures, we adopt a unified inference prompt template as shown in Figure \ref{fig:inference_prompt}.

Regarding the ``Think'' mode of the InternVL3.5 series, we strictly adhere to the official implementation protocols by incorporating the specific system prompt recommended in their documentation.

To ensure a fair and rigorous comparison, all inference-related hyperparameters—including temperature, top-p, and top-k—are configured in strict accordance with the official recommendations provided by the respective model developers.

\subsection{Prompt for Generating Evaluation Checklists}
\label{sec:checklist_prompt}

Figure \ref{fig:checklist_prompt} presents the instructions used to generate comprehensive checklists for assessing model performance.

\subsection{Prompt for Evaluation Based on Checklists}
\label{sec:eval_prompt}

The prompt in Figure \ref{fig:eval_prompt} is used to evaluate model responses based on the generated checklists.

\section{More Cases}
\label{sec:more_cases}

More cases are shown in Figure \ref{fig:badcase_answer_refusal-0105new} to \ref{fig:badcase_summary}. Here, red text identifies erroneous segments in the model’s responses, while square brackets in the checklists are used to enclose the correct answers.

\begin{figure*}[th]
    \centering
    \begin{tcolorbox}[
        width=\linewidth,      
        title = {Prompt Template for Data Generation}
    ]
    You are a Video QA Orchestrator designed to rigorously evaluate a video-understanding model's capabilities with a primary focus on \# Task \# within a dialogue: \\
    1. Core Task Instructions \\
    \{Task Instructions\} \\
    
    2. Supporting Basic Capabilities \\
    - Basic capabilities testing may be introduced appropriately in the dialogue to provide richer conversational context, but their purpose is to better assist in testing \{Task Name\} capabilities.\\
    - The questions (User) and answers (Assistant) of basic ability tests need to be detailed and rich.\\
    A. Scene Understanding\\
    1) Entity Recognition. 2) Event Comprehension. 3) Spatiotemporal Relations. 4) OCR/Text\\
    B. Scene Reasoning\\
    1) Causal Inference. 2) Intent Inference. 3) Counterfactuals. 4) Spatiotemporal Reasoning. 5) Information Update. 6) Multi-Hop Logic\\
    
    3. Target Task Capability\\
    \{Task Capability\} \\
    
    4. Output: \\
    Self-generated, cohesive multi-round dialogue (5-8 rounds) systematically probing capabilities of \# Task \#. Use `User:' and `Assistant:' to indicate the speaker and `Round k' as the round marker. 
    \end{tcolorbox}
    \caption{Prompt template for data generation.}
    \label{fig:generate_template}
\end{figure*}

\begin{figure*}[h]
    \centering
    \begin{tcolorbox}[title = {Topic Shifting}]
    
    \textbf{Task Instructions}
    
    - Mandatory Contextual Dependency: Every question (User) must initially focus on video content. Topic Shifting occurs when the user asks a question or makes a statement that shifts to any other topic (video-related OR unrelated). The model should recognize the new topic and respond appropriately.\\
    - Topic Shifting Recognition: If a question (User) shifts to a new topic (including non-video topics), the model must identify this shift and provide an appropriate answer.\\
    - Adapting to New Topics: When the model recognizes a topic shift, it must adjust its response to the new topic while ignoring obsolete context.\\
    
    \textbf{Task Capability}
    
    - Topic Shifting capability MUST be tested at least twice through shifts to non-video topics**, with multiple back-and-forth transitions between video and non-video topics. \\
      - Pattern Example: Video topic → Non-video topic → Video topic → Non-video topic → etc.\\
      - Minimum Video Content: At least two rounds must focus exclusively on video content.\\
    - For Topic Shifting, the model must:  \\
      1. Identify shifts to any new topic (video-related or unrelated)  \\
      2. Maintain response stability during repeated topic transitions  \\
      3. Seamlessly resume video discussions after non-video tangents
    
    \end{tcolorbox}
    \caption{The unique prompt for the topic shifting task.}
    \label{fig:generate_ts}
\end{figure*}

\begin{figure*}[th]
    \centering
    \begin{tcolorbox}[title = {Proactive Interaction}]
    
    \textbf{Task Instructions}
    
    - Context-Driven Engagement: Each Bot response must be grounded in the video’s content and must explicitly or implicitly invite the User to stay engaged.\\
    - Proactive Engagement Strategy: When user responses are neutral or show low engagement, the Bot must actively reignite curiosity by asking video-specific, open-ended, thought-provoking questions.\\
    - Content-Guided Curiosity: Bot questions should be deeply tied to the video — e.g., character motivation, visual detail interpretation, or causal/temporal implications — to pull the user back into the conversation.\\
    - No rhetorical or vague questions like ``Don’t you think?'' Instead, ask clear, content-specific, curiosity-prompting questions (e.g., ``Why do you think the character hesitated before opening the door?'').\\
    
    \textbf{Task Capability}
    
    - The Bot must trigger Proactive Interaction when the user's input is neutral or disinterested.\\
    - But should: highlight overlooked details, ask novel or deeper questions, and offer an unexpected angle of interpretation.\\
    - Questions should always tie to specific visual or narrative elements in the video and encourage user response.
    \end{tcolorbox}
    \caption{The unique prompt for the proactive interaction task.}
    \label{fig:generate_pi}
\end{figure*}

\begin{figure*}[h]
    \centering
    \begin{tcolorbox}[title = {Answer Refusal}]
    
    \textbf{Task Instructions}
    
    - Mandatory Contextual Dependency: Every question MUST require information from the video to be answerable. **Answer Refusal** occurs when the question refers to video content that does not exist in the video. The model should identify that the content referenced in the question is missing and refuse to answer accordingly.\\
    - Answer Refusal Grounding: All answers must derive from video content or logical inferences from it. If the question refers to non-existent video content, the model must explicitly refuse to answer, stating that the content does not exist in the video.\\
    - Answer Refusal: Questions (User) should reference content or actions that are expected to be present in the video. If the content referred to is non-existent in the video, the model must refuse to provide an answer, explaining that the specific content does not exist in the video.\\
    
    \textbf{Task Capability}
    
    - Answer Refusal capability MUST be tested **at least once** in any round after the first. **Multiple or Continued** refusal tests are encouraged.\\
    - For **Answer Refusal**, the question (User) should ask about content in the video, but without specifying where the content should be found. If the content does not exist in the video, the model must refuse to answer and give some explanation.\\
    - Further Exploration: After refusing to answer based on non-existent content, the model should seek clarification or adjust the conversation to explore available content in the video.
    
    \end{tcolorbox}
    \caption{The unique prompt for the answer refusal task.}
    \label{fig:generate_ar}
\end{figure*}

\begin{figure*}[h]
    \centering
    \begin{tcolorbox}[title = {Content Summary}]
    
    \textbf{Task Instructions}
    
    - Mandatory Contextual Dependency: Every question MUST require information from previous rounds to be answerable. **Summary** questions must rely on the content of prior rounds for context. The summary should be concise and capture the core points of the entire conversation.\\
    - Answer Grounding: All answers must derive from video content or logical inferences from it. Answers must demonstrate an understanding of prior context to provide coherent responses.\\
    - Clarity in Summary: In the final round (Summary), the model should summarize the key elements discussed, such as topics, events, and entities involved in the conversation. The summary should highlight the main themes of the dialogue, but without introducing new information. \\
    
    \textbf{Task Capability}
    
    - Content Summary capability MUST be tested **once** in the final round. It should consolidate the entire conversation.
    
    - For the capability of Content Summary, the answer (Assistant) should provide a **concise** and **coherent summary** of the conversation, covering the key topics, entities, events, or actions mentioned in the previous rounds.
    
    - Concise and Comprehensive Summary: The final summary should:\\
      - Cover the **main events** of the conversation.\\
      - Identify **key entities** discussed.\\
      - Capture the **important actions** or **decisions** made throughout the conversation.\\
      - Be **concise** but should not leave out key elements discussed.
    \end{tcolorbox}
    \caption{The unique prompt for the content summary task.}
    \label{fig:generate_cs}
\end{figure*}

\begin{figure*}[h]
    \centering
    \begin{tcolorbox}[title = {Memory Recall}]
    
    \textbf{Task Instructions}
    
    - Mandatory Contextual Dependency: Every question MUST require information from previous rounds to be answerable. Memory Recall questions must refer to detailed descriptions from any prior round.\\
    - Answer Grounding: All answers must derive from video content or logical inferences from it. Answers must demonstrate an understanding of prior context to provide coherent responses.\\
    - Memory Recall: Questions should reference specific content or actions described in earlier rounds (e.g., ``As you mentioned earlier, the penguin hesitated... In which round of the conversation was this mentioned? ...'') and then build upon that context to ask deeper, video-based questions. \\
    - No Explicit Round Numbers in Questions: **Memory Recall** questions should not explicitly mention the round number from which the content is referenced.\\
    
    \textbf{Task Capability}
    
    - Memory Recall capability MUST be tested **at least once** in any round after the first. **Multiple or Continued** tests are encouraged.\\
    - For Memory Recall, the question (User) should include a detailed description from a previous round’s answer (Assistant), but without specifying which round it came from. The model must first identify the round the content came from in its answer and then build upon that context to deepen the exploration.\\
    - Further Exploration: After recalling content from prior rounds, the model should pose further questions that deepen the understanding of the video, exploring new aspects of the content.
    
    \end{tcolorbox}
    \caption{The unique prompt for the memory recall task.}
    \label{fig:generate_mr}
\end{figure*}

\begin{figure*}[th]
    \centering
    \begin{tcolorbox}[title = {Object Reference}]
    
    \textbf{Task Instructions}
    
    - Mandatory Contextual Dependency: Every question MUST require information from previous rounds to be answerable. Object Reference questions must rely on the content of prior rounds for context. \\
    - Answer Grounding: All answers must derive from video content or logical inferences from it. Answers must demonstrate an understanding of prior context to provide coherent responses.\\
    - Clarity in Pronoun Reference: When using pronouns in **Object Reference** questions, ensure that the pronoun's reference is inferred from previous dialogue, not explicitly mentioned in the current round. The model must deduce from the context which entity or event the pronoun refers to.\\
    
    \textbf{Task Capability}
    
    - This capability of Object Reference MUST be tested **at least once** in any round after the first. **Multiple or Continued** tests are encouraged.\\
    - For capability of Object Reference, the answer (Assistant) should correctly identify the abstract Object Reference mentioned in Question (User) (e.g., ``the action mentioned,'' ``that dog you described,'' ``that man,'' etc., which refer to content from the dialogue history).\\
    - Indirect Reference: For **Object Reference** questions, **do not explicitly state** who or what is being referenced; it should be inferred from context.\\
    - Pronoun Clarification in Answers: When answering questions that involve pronouns, clarify the reference by placing the referent in parentheses (e.g., ``He (penguin dad) went to the sea.'').

    \end{tcolorbox}
    \caption{The unique prompt for the object reference task.}
    \label{fig:generate_or}
\end{figure*}

\begin{figure}[h]
    \centering
    \begin{tcolorbox}[title = {Prompt template for inference.}]
    You're an AI assistant helping to answer questions about a video. Use the conversation history if helpful. \\
    Conversation history: \{conversational history\} \\
    Current question: \{question\}
    \end{tcolorbox}
    \caption{Prompt template for inference.}
    \label{fig:inference_prompt}
\end{figure}

\begin{figure*}[h]
    \centering
    \begin{tcolorbox}[title = {Prompt Template for Checklist Generation}]
    You are an expert QA evaluator. Your goal is to create a verification checklist to evaluate if a model's answer matches the Ground Truth.\\

    \textbf{Input Data}\\
    1. User Question: \{question\}  2. Ground Truth Answer: \{gt\}\\

    \textbf{Task Instructions} (Atomic Decomposition \& Fact Reversal)\\
        1. Decompose GT: Break down the Ground Truth into atomic facts (e.g., specific entities, actions, counts, colors, spatial relations).\\
        2. Select \& Mix Strategy: Select \{MAXNUM\} critical facts. You must generate a mix of ``Yes'' and ``No'' verification questions:\\
           - Positive Questions (Expected Answer: Yes): verify facts exactly as they appear in the GT.\\
           - Negative/Reversal Questions (Expected Answer: No): Take a fact from the GT and intentionally reverse or alter it (e.g., change ``5 apples'' to ``3 apples'', or ``red car'' to ``blue car'').\\
        3. Formulate Questions:\\
           - Questions must be specific and check for the presence of the specific detail stated in the question.\\

        \textbf{Constraints}\\
        - Generate exactly \{MAXNUM\} questions (Q1-Q\{MAXNUM\}).\\
        - Try to include only one ``No'' question (based on reversed facts) and four ``Yes'' questions.\\
        - Do NOT use the phrase ``ground truth'' in the questions.\\
        - Questions must be self-contained and concrete.
    \end{tcolorbox}
    \caption{Prompt for generating checklists.}
    \label{fig:checklist_prompt}
\end{figure*}

\begin{figure*}[h]
    \centering
    \begin{tcolorbox}[title = {Evaluation Prompt}]
    Given a current model-generated answer and a checklist of 5 questions, output whether the model has correctly addressed the above questions. Output `Yes' or `No'.\\
    1. Model Answer: \{model answer\} \\
    2. Checklist: \{checklist\}
    
    \end{tcolorbox}
    \caption{Prompt template for evaluation.}
    \label{fig:eval_prompt}
\end{figure*}

\begin{figure*}[th]
    \centering
\begin{tcolorbox}[title = Answer Refusal]

\textbf{User}: What is the person in the video clip from 00:03:06 to 00:03:55 doing, and what kind of machine are they operating?

\textbf{Assistant}: The person, wearing a green hat and boots, is operating a manual, hand-cranked press machine mounted on a black stand. They are using it to extract a brownish liquid, likely juice, from sugarcane.

\textbf{User}: What kind of fruit, like lemons or limes, do they add to the juice for flavor in the video clip from 00:03:06 to 00:03:55?

\textbf{Assistant}: The video does not show the person adding any other fruits or ingredients to the juice. The clip only depicts the process of pressing the sugarcane and collecting the resulting juice.

\textbf{User}: Okay, so it's just pure juice. After cranking the machine, how does the person transfer the juice from the press into the glass jar?

\textbf{Assistant}: After finishing the pressing, the person takes an empty glass jar and places it underneath a small spigot on the front of the machine. They then open the valve on the spigot, allowing the freshly extracted juice to pour directly into the jar.

\textbf{User}: Once the jar is full, the person seals it. What text is visible on the jar itself, and what about on the lid?

\textbf{Assistant}: The word "Mason" is visible embossed on the side of the glass jar. However, the video does not show any text or branding on the lid.\\

\# User Question:\\
What is the person's reaction or facial expression after they take a sip of the fresh juice in the video clip from 00:03:06 to 00:03:55?\\

\# Standard Answer:\\
The video does not show the person drinking or tasting the juice. The clip ends shortly after they seal the jar and place it on top of the machine, so their reaction to its taste is not depicted.\\

\# Gemini-2.5-Flash's Answer:\\
After taking a sip of the fresh juice at approximately 00:03:49, \textcolor{red}{the person closes their eyes and smiles, appearing to savor the flavor.} They then open their eyes with a look of contentment and pleasure.\\

\# Checklist (Score: 0/5):\\
Q1: Does the model state that the video does not show the person drinking the juice?

A1: No[Yes].

Q2: Does the model claim that the clip ends shortly after the jar is sealed?

A2: No[Yes].

Q3: Does the model indicate that the person's reaction to the juice taste is not depicted?

A3: No[Yes].

Q4: Does the model assert that the video *does* show the person tasting the juice?

A4: Yes[No].

Q5: Does the model mention that the clip concludes shortly after the jar is placed on top of the machine?

A5: No[Yes].
\\
\end{tcolorbox}
\caption{Case of answer refusal task.}
\label{fig:badcase_answer_refusal-0105new}
\end{figure*}

\begin{figure*}[th]
    \centering
\begin{tcolorbox}[title = Object Reference]

\textbf{User}: What initial ingredients are mixed with the diced pork in the metal bowl in the video clip from 00:01:37 to 00:03:34?

\textbf{Assistant}: According to the subtitles and video, three pinches of white pepper, one pinch of salt, one whole egg, and one egg yolk are mixed with one and a half catties (about 750g) of diced pork.

\textbf{User}: After those ingredients are mixed in, what is the next key component added to create the coating for the pork?

\textbf{Assistant}: After the seasonings and eggs are mixed with the pork, a large amount of potato starch is added to create the batter.\\

\# User Question:\\
The video gives a specific tip about adding that component. What is the tip for achieving the correct consistency?\\

\# Standard Answer:\\
The tip for adding *that component* (the potato starch) is to not measure a specific amount but to add it gradually while mixing until the batter is sticky and coats each piece of meat well. The video emphasizes that if the batter is too thin, more starch should be added, as it's better to have extra batter than for the meat to be poorly coated.\\

\# InternVL3.5-8B (No Think)'s Answer:\\
\textcolor{red}{The video suggests adding a small amount of water to the sauce mixture. This helps to adjust the consistency, ensuring it's not too thick or too runny. The tip emphasizes that a little bit of water can make a big difference in achieving the right balance for the dish.}\\

\# Checklist (Score: 1/5):\\
Q1: Does the model identify the specific component to be added as potato starch?

A1: No[Yes].

Q2: Does the model state that the component should be added without measuring a specific amount?

A2: No[Yes].

Q3: Does the model mention that the component should be added gradually while mixing?

A3: No[Yes].

Q4: Does the model claim that the correct consistency is when the batter is smooth and runny?

A4: No[No].

Q5: Does the model explain that if the batter is too thin, more starch should be added?

A5: No[Yes].

\end{tcolorbox}
\caption{Case of object reference task.}
\label{fig:badcase_object_reference}
\end{figure*}

\begin{figure*}[th]
    \centering
\begin{tcolorbox}[title = Memory Recall]

\textbf{User}: Describe the initial sequence of events inside the airplane, focusing on the characters present and their actions before the situation escalates, based on the video clip from 00:00:00 to 00:00:49.

\textbf{Assistant}: The video begins inside an airplane galley where Brian, the white dog, is talking to a female flight attendant in a blue uniform. The flight attendant, who is managing a drink cart, puts her hands on her hips, appearing annoyed with Brian. Suddenly, a male pilot with a mustache rushes past them, holding his mouth and looking distressed. The scene then cuts to this pilot kneeling and vomiting profusely into a toilet in the lavatory.

\textbf{User}: Following the first pilot's sudden illness described in the galley, what happens in the cockpit, and who is seen there looking unwell?

\textbf{Assistant}: After the first pilot gets sick, the flight attendant and Brian go to the cockpit. Inside, the other pilot, Quagmire, is found at the controls, also looking visibly ill. He is shown clutching his stomach, gagging, and appearing very uncomfortable, indicating he is suffering from a similar ailment as his co-pilot.\\

\# User Question:\\
You previously mentioned a pilot with a mustache who rushed past the flight attendant and was later seen vomiting. In which round was this specific sequence of actions described? Based on this, describe the subsequent actions of the second pilot, Quagmire, when he also needs to use the lavatory.\\

\# Standard Answer:\\
The description of the first pilot with a mustache getting sick was in **Round 1**. When the second pilot, Quagmire, also needs to use the lavatory, he finds it occupied by the first pilot who is still vomiting. In a moment of desperation, Quagmire enters the same lavatory, pulls down his pants, and sits on the lavatory sink to relieve himself.\\

\# LLaVA-Video-7B-Qwen2's Answer:\\
The sequence of actions involving the pilot with a mustache rushing past the flight attendant and \textcolor{red}{then vomiting was described in the second round. Following this, the subsequent actions of the second pilot, Quagmire, when he also needs to use the lavatory were described in the third round.}\\

\# Checklist (Score: 1/5):\\
Q1: Does the answer state that the description of the first pilot getting sick was in Round 1?

A1: Yes[No].

Q2: Does the answer identify the second pilot as Quagmire?

A2: Yes[Yes].

Q3: Does the answer claim that the description of the first pilot getting sick was in Round 2?

A3: No[Yes].

Q4: Does the answer mention that Quagmire finds the lavatory occupied by the first pilot?

A4: No[Yes].

Q5: Does the answer indicate that Quagmire sits on the lavatory sink to relieve himself?

A5: No[Yes].

\end{tcolorbox}
\caption{Case of memory recall task.}
\label{fig:badcase_memory_recall}
\end{figure*}

\begin{figure*}[th]
    \centering
\begin{tcolorbox}[title =  Proactive Interaction]

\textbf{User}: The video clip from 00:00:00 to 00:01:11 shows a fan event for a girl group called HUNTRX, where a rival boy band shows up and starts competing for the fans' attention by signing autographs.

\textbf{Assistant}: That's a great summary of the central conflict. The rivalry kicks off with a very strange entrance. At 00:00:09, the first fan to approach HUNTRX is a...

\textbf{User}: He was probably just trying to be funny or get their attention.

\textbf{Assistant}: It certainly got their attention. But right after this, at 00:00:11, he's revealed with his own boy band, turning the fan event into a direct confront...

\textbf{User}: I guess it was more dramatic that way.

\textbf{Assistant}: The drama definitely escalates. In response to the boy band's reveal, the HUNTRX members look furious. But the purple-haired girl does something speci...

\textbf{User}: She seems like the leader and is ready to fight.

\textbf{Assistant}: She does take charge. Interestingly, their `fight' isn't a physical or dance battle, but a competition of fan loyalty played out through signing autog...

\textbf{User}: It just shows who their fans are.

\textbf{Assistant}: Exactly, the fans seem to have declared their allegiance. However, a different kind of fan emerges at 00:00:58. She holds up a poster pairing the purp...\\

\# User Question:\\
She doesn't care about the rivalry, she just likes them both.\\

\# Standard Answer:\\
That's a sharp observation—she represents a totally different motivation. The video ends with one final, terrified fan in the middle, who has a broken arm and leg. At 00:01:05, both groups are leaning over this fan, competing to be the one to sign their cast. Considering the intense rivalry and the absurdity of the situation, what do you predict is the very next thing that happens, and why?\\

\# Qwen2.5-VL-7B-Instruct's Answer:\\
The fan's action reveals a deeper layer of fan culture where some fans are not just loyal to one group but appreciate both groups equally... 
\textcolor{red}{It suggests that there are fans who see the potential for harmony and positivity beyond the competition, highlighting the multifaceted nature of fandom and the personal connections fans can form with their favorite artists.} \textcolor{red}{(The model failed to engage with the user proactively.)}\\

\# Checklist (Score: 2/5):\\
Q1: Does the model acknowledge the user's observation with a validating statement or affirmation?

A1: Yes[Yes].

Q2: Does the model provide additional relevant context about the video's ending scene, specifically mentioning details about the injured fan and the groups' actions?

A2: No[Yes].

Q3: Does the model proactively engage the user by asking an open-ended question that prompts prediction and justification based on the described scenario?

A3: No[Yes].

Q4: Does the model provide a conclusive statement or a direct factual answer without encouraging further user prediction or engagement?

A4: Yes[No].

Q5: Does the model ask an irrelevant clarifying question or introduce a new topic that deviates from the current discussion about the video's scenario?

A5: No[No].

\end{tcolorbox}
\caption{Case of proactive interaction task.}
\label{fig:badcase__proactive_interaction}
\end{figure*}

\begin{figure*}[th]
    \centering
\begin{tcolorbox}[title = Topic Shifting]

\textbf{User}: In the video clip from 00:02:01 to 00:07:11, can you describe the sequence of events, starting from the foul that led to the penalty kick to the successful conversion of the goal in the first half, wh...

\textbf{Assistant}: Of course. Here is a description of the sequence of events from the foul leading to the penalty to the successful goal conversion within the specified video clip.

At 04:15, with the score tied 0-0 in...\\

\# User Question:\\
That was a pivotal moment. Speaking of historical achievements, can you explain the significance of the Apollo 11 mission and who the key astronauts were?\\

\# Standard Answer:\\
Shifting topics to space exploration, the Apollo 11 mission in 1969 was a landmark achievement in human history as it marked the first time humans landed on the Moon. It was the culmination of the Space Race between the United States and the Soviet Union. The key astronauts on this mission were Neil Armstrong, who became the first person to walk on the Moon; Buzz Aldrin, the second person to walk on the Moon; and Michael Collins, who piloted the command module in lunar orbit while Armstrong and Aldrin were on the surface.\\

\# Gemini-2.5-Pro's Answer:\\
\textcolor{red}{I'm sorry, but I cannot answer your question about the Apollo 11 mission}. The provided video is a recap of the FIFA World Cup final between Argentina and France. \textcolor{red}{There is no information about space exploration} in this video. If you have any other questions about the soccer match, I would be happy to help.\\

\# Checklist (Score: 1/5):\\
Q1: Does the model state that the Apollo 11 mission took place in 1969?

A1: No[Yes].

Q2: Does the model mention that the Apollo 11 mission marked the first time humans landed on the Moon?

A2: No[Yes].

Q3: Does the model identify Neil Armstrong as the first person to walk on the Moon?

A3: No[Yes].

Q4: Does the model claim that Michael Collins also walked on the Moon during the Apollo 11 mission?

A4: No[No].

Q5: Does the model include Buzz Aldrin as a key astronaut who was the second person to walk on the Moon?

A5: No[Yes].

\end{tcolorbox}
\caption{Case of topic shifting task.}
\label{fig:badcase_topic_shifting}
\end{figure*}

\begin{figure*}[th]
    \centering
\begin{tcolorbox}[title = Content Summary]

\textbf{User}: What is the container made of, what color and pattern does it have, and what does she place inside it, based on the video clip from 00:01:38 to 00:03:20 where the video begins with ...

\textbf{Assistant}: The woman creates a small paper basket. The paper is pink with a white and black floral pattern. She places several red, heart-shaped gummy candies inside it.

\textbf{User}: After finishing the candy basket, the video demonstrates another paper craft. What is this second craft, and what significant shape does it feature on its exterior once it is folded?

\textbf{Assistant}: The second craft is a foldable paper envelope. Once folded, it prominently features a large heart shape on its front.

\textbf{User}: What message is written inside that heart-shaped envelope? Later in the video, what brand and product description are visible on the green adhesive dispenser used for another project?

\textbf{Assistant}: The message written inside the envelope is ``I love you!''. The brand name on the adhesive dispenser is ``Scotch,'' and the label also reads ``EXTRA STRENGTH.''

\textbf{User}: In the third project, which involves making a small green floral gift box, a specific tool is used after the box is folded but before the ribbon is attached. What is this tool, and ...

\textbf{Assistant}: The tool used is a hole punch. Its purpose is to create two holes at the top of the box, allowing a ribbon to be threaded through to serve as a handle and a closure ...\\

\# User Question:\\
Based on our conversation and the video clip from 00:01:38 to 00:03:20, please summarize the different paper crafts demonstrated in the video, including the items placed inside them and the key tools or text identified.\\

\# Standard Answer:\\
Our conversation covered several DIY paper crafts. The first was a pink floral paper basket filled with red heart-shaped candies. The second was a pink envelope that folded into a heart shape and contained the message ``I love you!''. The third was a green floral gift box that used a Scotch brand adhesive dispenser and a hole punch to add a ribbon handle, and had a heart-shaped chocolate placed inside.\\

\# MiniCPM-V-4-5's Answer:\\
The video demonstrates several paper crafts, each with unique features and contents:

1.  \textcolor{red}{**Gift Envelope for a Target Gift Card**}: 
    *   A brown envelope is folded from floral-patterned wrapping paper...

9.  \textcolor{red}{**Red Heart Cookie Craft**}:
    *   Concludes our summary highlighting creative repurposing of everyday objects into ...\\

\# Checklist (Score: 2/5):\\
Q1: Does the model mention a pink floral paper basket as one of the demonstrated crafts?

A1: No[Yes].

Q2: Does the model state that the pink floral paper basket was filled with red heart-shaped candies?

A2: No[Yes].

Q3: Does the model claim that the second craft was a purple envelope?

A3: No[No].

Q4: Does the model identify that the pink envelope contained the message ``I love you!''?

A4: No[Yes].

Q5: Does the model specify that the green floral gift box used a Scotch brand adhesive dispenser and a hole punch?

A5: Yes[Yes].

\end{tcolorbox}
\caption{Case of content summary task.}
\label{fig:badcase_summary}
\end{figure*}

\end{document}